\documentclass[preprint,12pt,3p]{elsarticle}

\usepackage{amssymb}

\usepackage{boxedminipage}
\usepackage{mathtools}
\usepackage{booktabs}

\usepackage{algpseudocode}
\usepackage{algorithm}

\usepackage{amsmath,enumitem}
\usepackage{subcaption}

\usepackage{xcolor}

\newcommand{\defeq}{\vcentcolon=}

\algnewcommand{\LineComment}[1]{\State \(\triangleright\) #1}

\DeclareMathOperator*{\argmax}{arg\,max}

\journal{Speech Communication}

\begin{document}

\begin{frontmatter}

\title{Transcribing Against Time}

\author[label1]{Matthias Sperber\corref{cor1}}
\cortext[cor1]{Corresponding author}
\ead{matthias.sperber@kit.edu}
\author[label2]{Graham Neubig}
\author[label1]{Jan Niehues}
\author[label3]{Satoshi Nakamura}
\author[label1]{Alex Waibel}
\address[label1]{Karlsruhe Institute of Technology, Germany}
\address[label2]{Carnegie Mellon University, USA}
\address[label3]{Nara Institute of Science and Technology, Japan}






\begin{abstract}

We investigate the problem of manually correcting errors from an automatic speech transcript in a cost-sensitive fashion. This is done by specifying a fixed time budget, and then automatically choosing location and size of segments for correction such that the number of corrected errors is maximized. The core components, as suggested by previous research \cite{Sperber2014}, are a utility model that estimates the number of errors in a particular segment, and a cost model that estimates annotation effort for the segment. In this work we propose a dynamic updating framework that allows for the training of cost models during the ongoing transcription process. This removes the need for transcriber enrollment prior to the actual transcription, and improves correction efficiency by allowing highly transcriber-adaptive cost modeling. We first confirm and analyze the improvements afforded by this method in a simulated study. We then conduct a realistic user study, observing efficiency improvements of 15\% relative on average, and 42\% for the participants who deviated most strongly from our initial, transcriber-agnostic cost model. Moreover, we find that our updating framework can capture dynamically changing factors, such as transcriber fatigue and topic familiarity, which we observe to have a large influence on the transcriber's working behavior.
\end{abstract}


\end{frontmatter}


\section{Introduction}
\label{sec:intro}

High quality speech transcripts are required in many different tasks, including for example web-based lecture archives, training data for automatic speech recognition (ASR), and input to downstream applications such as translation. Unfortunately, in realistic settings automatically created transcripts often contain too many errors to be useful as-is, and human annotators must be employed to improve their quality. This manual transcription is costly and time-consuming.

Previous works have attempted to improve the efficiency of manual supervision for speech transcription by dividing the speech into small segments that are convenient to transcribe \cite{Roy2009}, and choosing low-confidence segments of an ASR transcript that are more likely to contain errors \cite{Sanchez-cortina2012,Sperber2013}. Studies on cost-sensitive annotation have also shown that to maximize supervision efficiency, it is important to consider not only the number of errors that might be contained in a particular segment, but also the human supervision effort involved in correcting them \cite{Settles2008a,Tomanek2010a,Ramirez-loaiza2014a}. Recent work has shown that supervision efficiency can be further increased by training models to estimate the transcriber effort and potential error reduction of each segment, and then explicitly optimizing the location of segment boundaries \cite{Sperber2014}.

However, this use of models to predict transcriber effort has a downside: these models need to be trained. Thus, before starting the main transcription task, it is necessary to perform enrollment, where the transcriber annotates a certain amount of randomly selected enrollment data, which is then used to train the predictive cost models. However, enrollments are time-consuming, costly, and may be impractical; for example, in a crowd sourcing situation. In addition, these predictive models remain static and cannot account for dynamically changing factors such as the annotator's topic familiarity, fatigue, or increasing experience.
 
In this work, we introduce a framework that removes this need for enrollment by updating cost models and corresponding choice of data to annotate on-the-fly (during the ongoing transcription process).\footnote{We have previously presented the basic idea of this method and verified it through simulation in the proceedings of SLT2014 \cite{Sperber2014a}. In this paper, we expand the description and add a full user study with 12 expert and non-expert participants.} This framework allows us to work within a fixed time limit for manual correction of a transcript or a series of transcripts. We first start with a general, initial cost model used to compute a first selection of segments for correction. During the ongoing transcription process, the cost model is gradually improved and adapted toward the particular transcriber, and the choice of segments is updated to reflect both the updated cost model and the actual remaining time at a particular point. Locations and lengths of segments to annotate are chosen to optimize annotation efficiency, as proposed in previous work \cite{Sperber2014}.

A number of challenges arise in this proposed dynamic annotation framework. (1) A suitable time limit must be chosen and incorporated into our framework as a stopping criterion. (2) Annotation effort is difficult to predict, because there are large differences not only between transcribers, but also for a single transcriber between different segments. Accurate cost modeling is important both for selecting suitable segments that are efficient to annotate, and for the amount of selected segments to be appropriate such that the time budget is not over- or underspent. (3) Choosing segments to annotate is a computationally difficult problem, but needs to be  done quickly, as we desire to update the segmentation during the ongoing transcription process.

Our solutions to these challenges are as follows:
(1) We propose to limit the transcription time budget to a fixed value, in order to reflect one's desired cost-quality tradeoff. How to choose a suitable time budget is task specific, but we argue that it is more practical than configuring a confidence threshold as was required in some previous works \cite{Sanchez-cortina2012,Sperber2013,ValorMiro2015}. Moreover, by periodically updating segmentations ($\mathsection$\ref{sec:framework}) we can recover from inaccurately predicted correction times, making sure that the budget is not over- or underspent.
(2) We propose an approach of starting with an initial, general cost model and gradually adapting it toward the particular transcriber ($\mathsection$\ref{sec:cost_utility}). This has the potential to remove the need for enrollment, while being able to model transcriber-specific as well as dynamically changing characteristics.
(3) We propose a new, more efficient algorithm for choosing which segments of the ASR transcript to annotate using the penalty method ($\mathsection$\ref{sec:algo}). We demonstrate this algorithm to be fast enough to update segmentations during the transcription process, without degrading quality compared to the original, much slower algorithm \cite{Sperber2014}. 

We first conduct a partly simulated evaluation approach ($\mathsection$\ref{sec:simulations}). Results show that our method outperforms both cost-insensitive baselines and cost-sensitive baselines without updates. An analysis shows that efficiency gains are attributed to (1) increasingly accurate cost models, (2) adjusting to the actual remaining time budget with each update, and (3) choosing a sensible (although crude) initial cost model that includes cognitive overhead. Finally, we conduct a realistic user study ($\mathsection$\ref{sec:user_study}) in a typical scenario where several previously unknown transcribers conduct only a limited amount of work. We notice large differences between different transcribers, supporting our claim that transcriber-specific modeling is crucial. Moreover, we observe relative productivity gains of 15\% on average, and 42\% for those participants who deviated most from the initial cost model. The gains of using our updating framework were especially strong in the case where the initial ASR transcript was already of relatively high quality. 

\section{Related Work}
Adaptive user models have been studied in the human computer interaction community \cite{Zigoris2006}, with very different requirements from ours. We are not aware of previous works on adaptive user modeling or choice of data to annotate in the context of cost-sensitive annotation or computational linguistics in general. A closely related work on quality estimation \cite{deSouza2015} adapts an automatic quality estimator online and in a multi-task fashion, as more and more in-domain samples are annotated over time.

Efficient supervision strategies have been studied across a variety of NLP-related research areas, and received increasing attention in recent years. Examples include post editing for speech recognition \cite{Sanchez-cortina2012}, interactive machine translation \cite{Gonzalez-Rubio2010}, active learning for machine translation \cite{Haffari2009,Gonzalez-Rubio2011} and many other NLP tasks \cite{Olsson2009a}, to name but a few studies. Most of these do not model annotation cost explicitly. However, it has been recognized that correcting only the instances of highest utility is often not optimal in terms of efficiency, since these parts tend to be the most difficult to manually annotate \cite{Settles2008a,Miura2016}. As a solution, the idea of using an annotator cost model to predict the supervision effort has been developed \cite{Settles2008a,Tomanek2010,Specia2011,Cohn2013,Sperber2014}, which inspired our approach as well. Note that these previous works estimate static, annotator-specific cost models via enrollment, whereas our dynamic approach does not require enrollment.

Some studies have addressed the problem of balancing utility and cost in the context of active learning. A greedy approach to combine both into one measure is the ``bang-for-the-buck'' approach \cite{Settles2008a}, where utility divided by effort is used as a per-instance efficiency measure. Such an approach can be effective for selecting isolated instances for annotation, but is problematic when selecting segments that can overlap and conflict with one another, as in our task. A more theoretically founded scalar optimization objective is the net benefit (utility minus costs) as proposed by \cite{Vijayanarasimhan2009}, but unfortunately is restricted to applications where both can be expressed in terms of the same monetary unit. \cite{Vijayanarasimhan2010} and \cite{Donmez2008} use a more practical approach that specifies a constrained optimization problem by allowing only a limited time budget for supervision, similar to our approach. Note that works on annotation apart from the active learning community have usually assumed stopping criterions based on confidence thresholds \cite{Sanchez-cortina2012,Sperber2013,ValorMiro2015}, which measure only relative improvement and may not be intuitive to configure in practice.

\section{Background: Static Segmentation}

Our advocated ``transcribing against time'' paradigm builds upon the following three lines of work:
\begin{itemize}
\item Roy et al.\ \cite{Roy2009} argue that human effort should be spent on transcription of speech, not its segmentation. Segmentation is time-consuming if performed manually, and can be done reliably in an automatic fashion. Segmentation is also a very important step, because suitable segment size leads to increased transcription efficiency.
\item Rodriguez et al.\ \cite{Rodriguez2007} propose the computer-assisted transcription (CAT) approach, in which a transcription is first created automatically, and then corrected manually where necessary. In their approach, a human checks the complete transcript and performs corrections whenever errors are found.\footnote{CAT also improves the automatic transcript on-the-fly using the transcriber's corrections. This is not the focus of our work, but could be integrated into our proposed updating framework.}
\item Sperber et al.\ \cite{Sperber2014} argue that the decision which segments to correct and which not, should also be done automatically to reduce human effort. They show how to select segment locations and sizes according to a desired utility-cost tradeoff, by exploiting confidence scores and a transcription cost model (CM). They report 25\% time savings compared to a baseline that selects segments using only confidence scores but no CM. 

\end{itemize}

\subsection{Constrained Optimization Problem}
\label{sec:optimization_problem}

\begin{figure}[h]
	\begin{boxedminipage}{0.98\textwidth}
	\begin{center}
\includegraphics[width=5.0in]{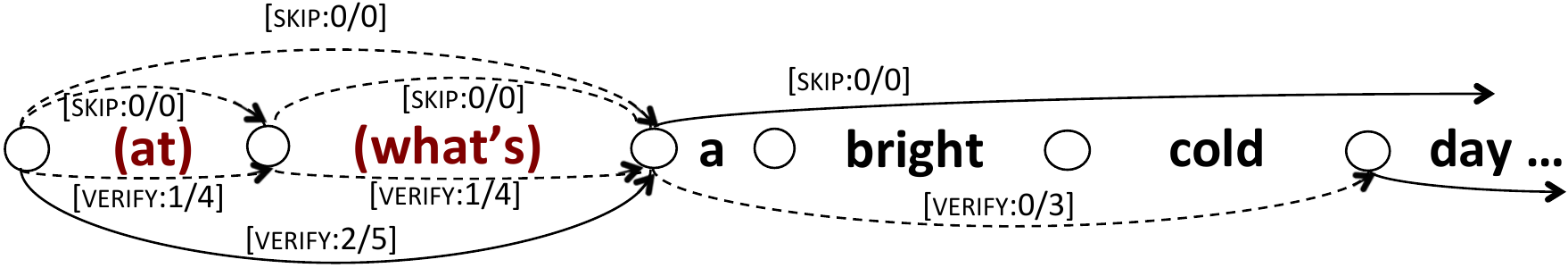}
	\end{center}
\end{boxedminipage}
	\caption{Illustration of the segmentation problem. A graph of different possible segmentations for the ASR transcript of the sentence fragment \emph{``It was a bright cold day \dots''} is shown. Edges are segments (some edges are omitted for readability), and nodes potential segment boundaries. Segments are labeled with [supervision mode:predicted utility/predicted cost]. For optimal supervision efficiency, the solid edges might be preferable over the dashed ones.}
	\label{fig:segmentation_problem_illustration}
\end{figure}

In particular, our work extends the optimization framework \textsc{Sesla} (short for Segmentation for Efficient Supervised Language Annotation), as proposed in \cite{Sperber2014}. Given an ASR transcript to be corrected by a human annotator, \textsc{Sesla} determines a segmentation of the transcript, and makes a decision whether or not to transcribe each segment, such that annotation efficiency is optimized. The idea is that we naturally want to concentrate on annotating parts that yield high utility, in terms of number of errors\footnote{By number of errors, we mean Levenshtein distance to a reference transcript.} that are removed by supervising them. While choosing very small segments (e.g.\ single words) would allow us to concentrate only on parts of maximum utility, longer segments are desirable from a cognitive point of view as they reduce cognitive overhead due to switching between different parts of the transcript. \textsc{Sesla} attempts to balance both cost and utility.

Formally, \textsc{Sesla} searches for a segmentation of the $N$ words of an ASR transcript $w_1^N$ into $M\le N$ segments with boundaries at $s_1^{M+1}=(s_1{=}1,s_2,\dots,s_{M+1}{=}N+1)$. A segment boundary marker $s_i$ is interpreted as placing a segment boundary before the $s_i$-th word (or the end-of-transcript marker for $s_{M+1}{=}N+1$). Each segment is further associated with a supervision mode $m_j\in K$. In this work these supervision modes are either \textsc{Verify} (the segment should be verified and potentially corrected), or \textsc{Skip} (the segment will not be verified). This segmentation is selected based on predictive models that estimate supervision cost (here: correction time) and utility (here: number of corrected errors) for any particular segment. For each segment $w_a^b$ and supervision mode $k$, we denote utility by $\operatorname{util}_k(w_a^b)$ and cost by $\operatorname{cost}_k(w_a^b)$. Note that cost and utility for the \textsc{Skip} mode are always 0. Figure~\ref{fig:segmentation_problem_illustration} illustrates the segmentation problem.

We desire a segmentation that maximizes the total utility, while keeping the total cost within a cost budget $T$:
\begin{align}
\argmax_{M;s_1^{M+1};m_1^M}\;\;\;\;\;&{\sum_{j=1}^M\left[\operatorname{util}_{m_j}\left(w_{s_j}^{s_{j+1}}\right)\right]}\label{eq:optimization_objective} \\
\text{subject to } \;\;\;\;\;& \sum_{j=1}^M\left[\operatorname{cost}_{m_j}\left(w_{s_j}^{s_{j+1}}\right)\right]\le T\text{  } & \label{eq:optimization_constraint}
\end{align}

We will introduce an efficient method for solving this optimization problem in $\mathsection$\ref{sec:algo}. The CM is a transcriber-dependent regression model, while the utility model (UM) is transcriber-independent and based on ASR confidences. Both will be described in $\mathsection$\ref{sec:cost_utility}.

\section{Proposed Dynamic Updating Framework}
\label{sec:framework}

\textsc{Sesla} as proposed previously produces a static segmentation once before starting the transcription. It relies on a CM that is trained via an enrollment, which is costly and takes up time that the transcriber might otherwise have invested in being productive. For instance, in \cite{Sperber2014} an enrollment of roughly 30 minutes is carried out. While investing enrollment time will improve supervision efficiency and pay off in the long run, this approach may not be economical at all for transcribers that supervise only a small amount of data, as is common in crowd-sourcing situations.

In this work, we improve upon this situation by allowing transcribers to be productive from the start. We create an initial segmentation with a crude initial CM, and iteratively improve the CM and consequently the segmentation as the transcription is underway. There are several advantages to this approach: (1)~The transcriber enrollment is removed, instead the transcriber performs useful annotation from the start. (2)~The segmentation will grow increasingly efficient due to the improving CM, starting out rather crude in the beginning, and eventually reaching or surpassing the efficiency that would have been reached using a CM trained via enrollment. (3)~Each updated segmentation will take into account the actual remaining time budget, which may differ from what had been predicted as remaining when the transcriber would reach a certain position in the ASR transcript. This solves an issue pointed out in the previous work \cite{Sperber2014}, in which systematic errors in time predictions resulted in considerable over- or underspending of the given time budget. Re-segmenting ensures that the remaining time is used optimally, for example by skipping some of the less promising segments if the remaining time budget had been overestimated.

\begin{figure}[tb]
	\begin{center}
\includegraphics[width=0.7\linewidth]{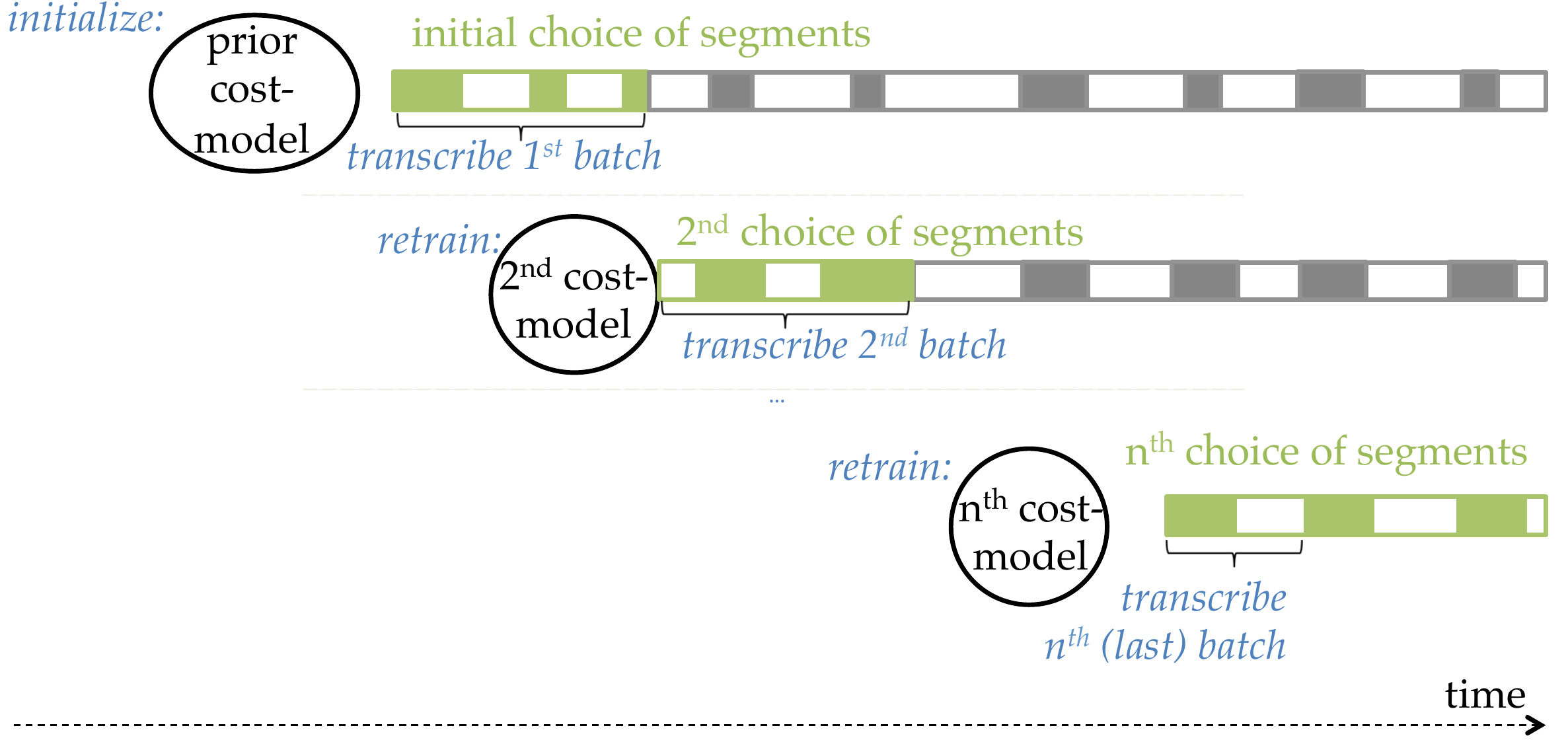}
	\end{center}
	\caption{Proposed updating framework. An initial choice of segments is determined based on an initial, transcriber-independent CM. Periodically, the CM is adapted based on the observed annotations, and the choice of segments is updated to reflect the new CM and the actual remaining time budget.}
	\label{fig:updating_framework_diagram}
\end{figure}

Figure~\ref{fig:updating_framework_diagram} illustrates the updating approach. An initial CM is used to create an initial segmentation. The annotator starts transcribing the resulting segments in temporal order. Up unto this point, the process is identical to the static method except for how the CM is obtained. However, unlike with the static method, only the first few segments are transcribed. After that, the CM is updated based on the observed annotation times, and the remainder of the initial transcript (everything that comes {\it after} the last annotated segment) is re-segmented to reflect the updated CM and the actual remaining time budget. This cycle of transcribing and updating repeats until the last segment has been reached. Because segmentation updates reflect the actually remaining time budget at the point of each update, the end of the transcript can be expected to be reached very closely to the point at which the time budget is exhausted. It is important to note that we follow the temporal order of the segments in the transcription process. This is a desirable property, as in our experience jumping forwards and backwards through segments makes it difficult for the transcriber to grasp the content of the speech and may lead to transcription errors.

\begin{algorithm}
\caption{Updating Annotation Framework}
\label{algorithm1}
\begin{algorithmic}
\LineComment{given: $T$ (time budget), $w_1^N$ (ASR transcript), $B$ (batch size [annotation time per batch in seconds]), $M_U$ (UM, fixed)}
\State $M_C\gets M_{C}^{(0)}$\Comment{initialize cost model $\mathsection$\ref{sec:cost_model}}
\State $j\gets 1$\Comment{set transcription position to first word}\\
\Call{Segment}{$w_1^N, M_C, M_U, T$}\Comment{solve optimization problem $\mathsection$\ref{sec:optimization_problem}}
\While{$t_{\text{remain}}>0$}

	\Call{SuperviseBatch}{$w_j^N, \min(B, t_{\text{remain}})$} \Comment{actual transcription}
	
	$j\gets$ resulting transcription position
	
	\Call{Update}{$M_C$, newly observed times} \Comment{$\mathsection$\ref{sec:cost_model}}

	\Call{Segment}{$w_j^N, M_C, M_U, t_{\text{remain}}$}\Comment{solve optimization problem $\mathsection$\ref{sec:optimization_problem}}
\EndWhile
\end{algorithmic}
\end{algorithm}
Formally, we propose to update segmentations as in Algorithm~\ref{algorithm1}. We initialize the cost model $M_C$ such that it will rely solely on a prior distribution to be defined. The transcription position $j$ is initialized to the first word, and an initial segmentation of the complete ASR transcript is created. The transcriber starts transcribing the first batch of segments from the current position $j$, until $B$ seconds have passed, at which time $j$ is updated to the new position at which transcription had stopped. The cost model $M_C$ is updated using the observed supervision times.\footnote{Note that the UM remains unchanged.} Finally, the remainder of the  transcript that has not yet been supervised ($w_j^N$) is re-segmented using the updated CM and the remaining time budget $t_\text{remain}$, and another batch of $B$ seconds is transcribed. Transcription is stopped when the time budget is exhausted, or the end of the transcript has been reached.

\section{Predicting Cost and Utility}
\label{sec:cost_utility}

\subsection{Cost Model}
\label{sec:cost_model}
As a cost model, a regression model that predicts transcription time is learned in a supervised fashion from observed transcriptions. With the original, static method, CM training examples were collected via transcriber enrollment, whereas we will show how to use examples collected on-the-fly. We employ Gaussian Process (GP) regression \cite{Rasmussen2006}, a Bayesian technique for which state-of-the-art performance has been reported, especially in the case of relatively low dimensionality and limited training data \cite{Cohn2013,Specia2013}. We intrinsically confirmed its superiority over linear regression and support vector regression in preliminary experiments. As an additional advantage, it allows convenient specification of a prior distribution over transcription costs, a feature that we will exploit to bootstrap our CM. Input features are the segment length (number of words), audio duration (seconds), and average word-level confidence. We expect length and duration to increase transcription cost, while low confidence words are potentially harder and more expensive to transcribe.

To create the initial segmentation, we need a reasonable initial CM. Since transcribed data from similar users or tasks may not be available, we rely solely on a manually defined prior inspired by our intuition about the transcription task. We argue that to create a sensible initial segmentation, the CM can be crude, but should capture two key observations, namely that longer segments take longer to supervise, and that the transcriber will need some time to process the context switch for each new segment. We therefore specify the prior cost as $2+n$ seconds, where $n$ is the segment length. Considering our transcript correction task, this prior underestimates the actual cost, but we deliberately avoid hand-tuning the prior to simulate a situation where little task expertise is available.\footnote{If possible, it is preferable to specify an underestimating prior rather than an overestimating prior, as we will point out in $\mathsection$\ref{sec:user_study}.}

Updates are performed by adding the new examples to the training data and retraining the CM. As more and more examples are collected, the trained model will move farther away from the prior distribution. For efficiency reasons, we use only the most recent 1000 examples for retraining.

\subsection{Utility Model}
Utility, defined as the numbers of substitution, insertion, and deletion errors removed from the transcript, is modeled in a transcriber-agnostic fashion. Here, we simply approximate segment utility as $\operatorname{util}\big(w_{s_j}^{s_{j+1}}\big)=1-\frac{1}{s_{j+1}-s_j}\sum_{i{=}s_j}^{s_{j+1}-1}c_i$, where $c_i$ is a scaled word confidence score for the $i$-th word, retrieved from the ASR. Note that strictly speaking, our approximation does not capture deletion errors. In this work, we do not update the UM dynamically. While it may be conceivable to learn an average edit distance on-the-fly for each transcriber, this would only scale the optimization objective (utility) by a constant factor, without changing the resulting segmentation.

\section{Computing Segmentations via the Penalty Method}
\label{sec:algo}

To make on-the-fly updates of the segmentation practical, the optimization problem (Equations \ref{eq:optimization_objective} and \ref{eq:optimization_constraint}) must be solved rapidly so as not to incur waiting times while updating. The previous work \cite{Sperber2014} used a general-purpose integer linear program (ILP) solver to find an approximate solution, but this approach is not sufficiently fast for updating segmentations during the transcription process. In this section, we introduce a new segmentation algorithm based on the well-known penalty method \cite{Fiacco1968} that we will demonstrate to be dramatically more time- and memory-efficient ($\mathsection$\ref{sec:segmentation_algorithms}).

Note that our optimization problem is reminiscent of other segmentation problems that are often solvable by dynamic programming (DP) in polynomial time. Unfortunately, the global constraint in Equation~\ref{eq:optimization_constraint} that enforces the time budget prevents us from using DP.
Our strategy for solving the problem is to replace the constrained objective function by an unconstrained penalty function $s(\cdot)$:
\begin{align}
\argmax_{M;s_1^{M+1};m_1^M}{\sum_{j=1}^M\left[s_{\lambda;m_j}\left(s_j,s_{j+1}\right)\right]}, \label{eq:unconstrained_objective} \\
\text{where } s_{\lambda;k}(i,j)\defeq \operatorname{util}_{k}(w_{i}^{j})-\lambda \operatorname{cost}_{k}(w_{i}^{j}). \label{eq:penalty_function}
\end{align}

For a fixed value for the penalty parameter $\lambda$, the optimization problem in Equation~\ref{eq:unconstrained_objective} is solvable via DP as we will soon show. By varying the value of $\lambda$, we can obtain different solutions for which we cannot increase their utility without increasing cost and vice versa, which are also called {\it Pareto-optimal} segmentations. Note that the optimal solution to the constrained problem corresponds to one of the Pareto-optimal solutions of the new penalized formulation. Hence the search problem reduces to finding the optimal value for $\lambda$. 

Intuitively, if a given $\lambda$ results in a Pareto-optimal segmentation that overspends our time budget (we say this segmentation is {\it infeasible}), we should increase $\lambda$, thus penalizing costly segments more strongly. Similarly, if the budget is underspent (the segmentation is {\it feasible}), $\lambda$ should be decreased. We follow this simple intuition by iteratively first computing Pareto-optimal segmentations, and then adjusting $\lambda$. While finding the optimal value for $\lambda$ is NP-hard, we can efficiently find an arbitrarily good approximation via a binary search.

\begin{algorithm}
\caption{Iterative Penalized Dynamic Programming}
\label{algorithm2}
\begin{algorithmic}
\LineComment{Initialize lower-/upper-bound segmentations $S_L,S_U$ and penalties $\lambda_L,\lambda_U$}
\While{$\operatorname{util}(S_U)/\operatorname{util}(S_L)>1+\epsilon$}

	
	$\lambda'\gets(\lambda_U+\lambda_L)/2$
	
	$S'\gets$\Call{SegmentDP}{$\lambda'$}

	\If{$\operatorname{cost}(S_U)\ge \operatorname{cost}(S')\ge T$}
	
		$\lambda_U,S_U\gets\lambda',S'$
	
	\Else \Comment{i.e., $\operatorname{cost}(S_L)\le \operatorname{cost}(S')\le T$}
		
			$\lambda_L,S_L\gets\lambda',S'$
		
	\EndIf
\EndWhile
\end{algorithmic}
\end{algorithm}

The segmentation algorithm is outlined in Algorithm~\ref{algorithm2}. We keep track of an upper bound and a lower bound segmentation $S_U,S_L$, with corresponding penalties $\lambda_U,\lambda_L$. The upper bound refers to the lowest-scoring infeasible segmentation seen so far, and the lower bound is the highest-scoring feasible segmentation so far. For initialization, we first try an arbitrary value for the penalty and assign it to either $\lambda_L$ or $\lambda_U$, depending on whether the segmentation was feasible or not. To find the missing value for the other parameter $\lambda_U$ ($\lambda_L$), we then repeatedly multiply (divide) $\lambda$ by a constant factor (here: $10$), until an (in-)feasible Pareto-optimal solution is found.

In the main loop of the algorithm, we consider a new penalty $\lambda'$, halfway between the upper- and lower-bound values. We compute a corresponding Pareto-optimal segmentation $S'$ via DP, and update the lower or upper bound, depending on whether the segmentation was feasible or not. The loop stops when the gap between upper and lower bound utilities is below a threshold $\epsilon$.

The DP algorithm that finds Pareto-optimal segmentations, given a penalty $\lambda$, computes the total penalized score $a_{j}$ of the best segmentation of $w_{1}^{j}$, and keeps back pointers to find an optimal segmentation as follows:

\begin{eqnarray*}
a_{1} & = & \max_{k\in K}s_{\lambda;k}(1,1)\\
a_{j} & = & \max_{i<j}(a_{i}+\max_{k\in K}s_{\lambda;k}(i,j))
\end{eqnarray*}

The DP has computational complexity $O(N^2)$. By limiting the segment size to be at most $R$ (here: 20), the complexity reduces to $O(RN)$. Moreover, the number of iterations of the binary search for $\lambda$ depends on the approximation threshold $\epsilon$, but generally does not depend on $N$, so the overall algorithm's complexity is essentially linear in the length of the ASR transcript. In practice, we observed convergence after typically less than 20 iterations.

\section{Simulated Experiments}
\label{sec:simulations}

To be able to compare a large number of different settings, we first conducted partly simulated experiments. We asked a real, non-expert transcriber to transcribe in chronological order 200 segments randomly chosen from the evaluation data, balanced across different lengths and average confidence scores. We measured the time taken to transcribe each segment. This enrollment process took about one hour. Based on this data, we trained an {\it oracle time model} and assumed that the human transcriber behaves according to this oracle time model. We further assumed that the transcriber successfully transforms the ASR transcript of every supervised segment into the corresponding reference transcript, without making mistakes. As a specific method to train this model we trained a GP regressor, and used this model as the gold standard that the tested CMs will try to reproduce. To ensure that the oracle model was sufficiently different from the initial CMs to be adapted, we applied a different prior for the oracle model. This prior was set to the least-squares linear regression fit on the same data. We then distorted the oracle model's predictions with multiplicative gamma-distributed noise, to serve as our final oracle time model. The multiplicative noise had a mean of $1$ and moderate variance of $0.01$. The batch size $B$ is set to 2.5 minutes, unless otherwise noted, i.e.\ the proposed updates are performed every 2.5 minutes.\footnote{We expect and will confirm low batch sizes (frequent updates) to be better. 2.5 minutes is the lowest value we tried in this set of experiments.} All results are averaged over 10 simulation runs, with the order of talks and noise multipliers chosen at random.

For the simulations, we considered a correction scenario where the transcriber is given an ASR transcription with the goal to remove as many errors as possible, given a 100 minute time budget. As transcription data, we used 10 TED talks\footnote{www.ted.com} (short presentations by skilled speakers, the total length was 104 minutes or 17.8k words). We concatenated the 10 TED talks so that the budget is allocated jointly over all talks. The erroneous transcripts were created using the Janus speech recognition toolkit \cite{Soltau2001} with a simple TED-optimized setup. The word error rate on our test set was 22.3\%, with a total of 3978 errors. The reference transcripts were carefully transcribed and of high accuracy. Our CMs are parametrized similar to \cite{Sperber2014}: We employ scaled lattice posterior probabilities as our UM, and GP regression with a squared exponential kernel as our CM, predicting log times to avoid negative values.\footnote{GP regression was done using GPy: github.com/SheffieldML/GPy} Noise and kernel variances for the GP regressors were set to $\log(5 \text{ sec})$. We empirically confirmed that these parameters are sensible, but did not fine-tune them because in a practical situation, parameter tuning data might not be available.

We configured the proposed segmentation algorithm to find a solution within 1\% of the optimal solution, and restricted the maximum segment length to 20 words. The average resulting segment length was 8 words or 3 seconds.

\subsection{Segmentation Algorithms}
\label{sec:segmentation_algorithms}
\begin{figure}[tb]
	\begin{center}
\includegraphics[width=0.5\linewidth]{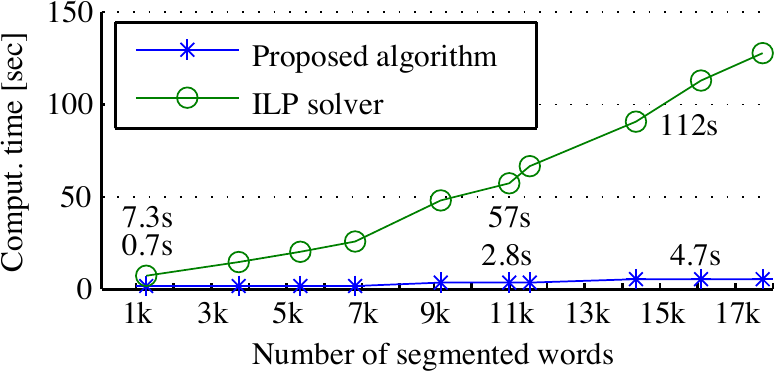}
	\end{center}
	\caption{Computation time needed to segment different amounts of data using an ILP solver or our proposed algorithm.}
	\label{fig:algo_compare}
\end{figure}

We first compared the computational efficiency of segmenting according to the algorithm proposed in $\mathsection$\ref{sec:algo}, as opposed to the baseline method of formulating the problem as an integer linear program and using an off-the-shelve ILP solver. As our solver, we used GUROBI, which is highly optimized commercial software and among the fastest ILP solvers available,\footnote{See benchmark on http://plato.asu.edu/ftp/milpc.html (accessed June 6, 2016)} while our Java implementation of the proposed algorithm is straightforward, with little code level optimization. We used the initial CM to segment increasingly large subsets of our data. Both algorithms were run on a single processor, and stopped after producing a solution within 1\% of their respective upper bounds. Figure \ref{fig:algo_compare} shows the results. It can be seen that the proposed method is considerably faster. It stays within a computation time of around 3 seconds, while the ILP solver needs more than two minutes to segment all data. Further, the proposed method's memory consumption grows only linearly in the number of words, needed to store the DP scores. In contrast, for segmenting the complete dataset, we observed memory consumption one or even two orders of magnitude higher for the ILP solver. We use the proposed algorithm throughout the following experiments.

\subsection{End-to-end Effectiveness of Updating Framework}

\begin{figure}[tb]
	\begin{center}
\includegraphics[width=\linewidth]{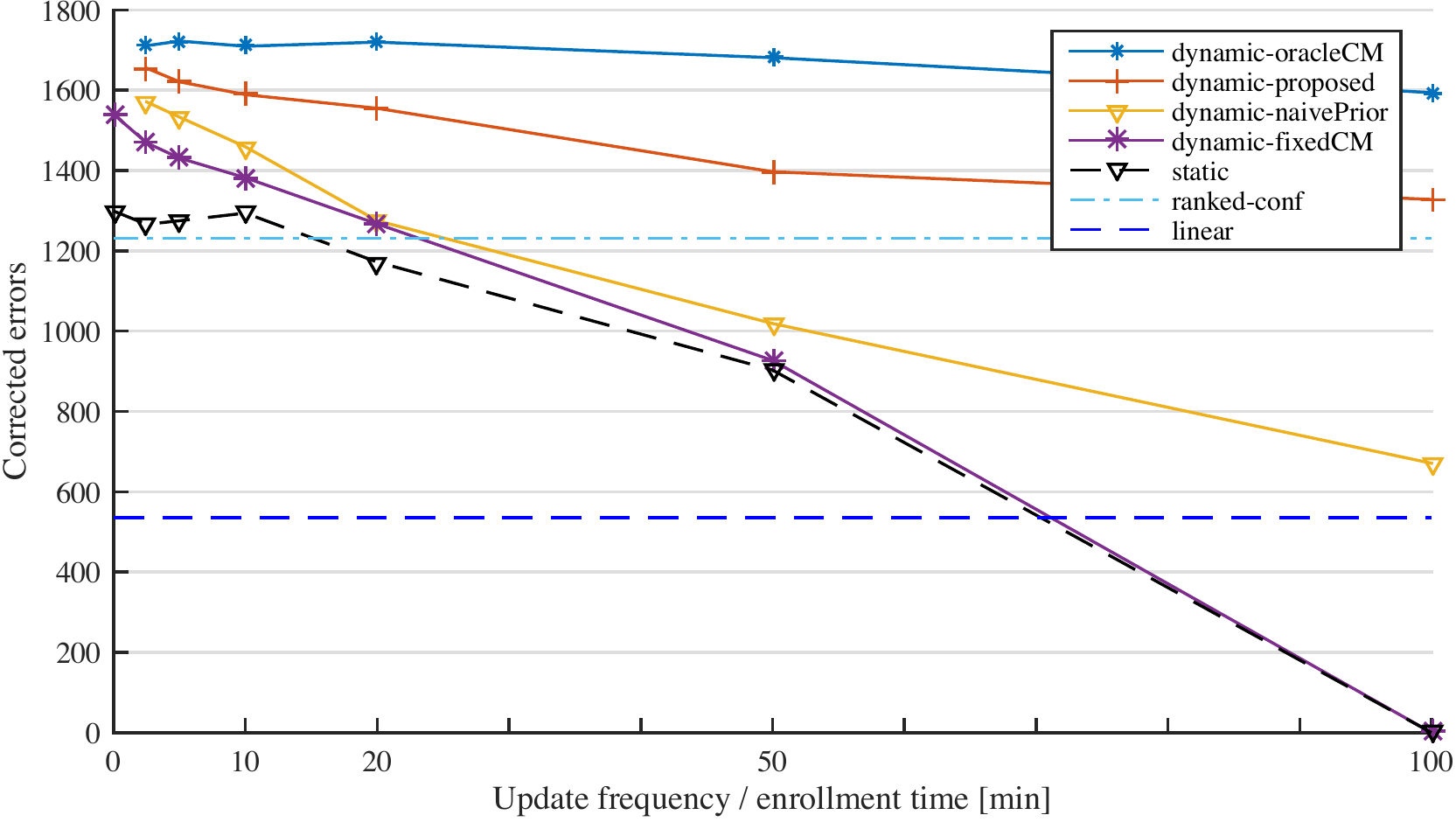}
	\end{center}
	\caption{Errors corrected for \textsl{dynamic-proposed} and several other settings (see Table~\ref{tab:settings_overview}), plotted against update frequency (first 3 methods) or enrollment time (following 2 methods) as described in the text.}
	\label{fig:batches}
\end{figure}

In this section, we evaluate the end-to-end effectiveness of the proposed updating framework, compared to several baseline approaches, in terms of the final number of errors removed after 100 minutes of transcription time. As Figure \ref{fig:batches} indicates, with the \textsl{dynamic-proposed} setup 1655 errors are removed when performing updates every 2.5 minutes, with the utility decreasing for fewer updates, until only 1328 errors are removed when performing no updates to the initial segmentation. 

\begin{table}[tb]
\centering
\label{my-label}
\begin{tabular}{@{}lcccc@{}}
\toprule
                & \begin{tabular}[c]{@{}l@{}}utility\\ model\end{tabular} & \begin{tabular}[c]{@{}l@{}}cost\\ model\end{tabular} & \begin{tabular}[c]{@{}l@{}}cost prior models\\ segment overhead\end{tabular} & \begin{tabular}[c]{@{}l@{}}segmentation\\ updates\end{tabular} \\ \midrule
\textsl{dynamic-oracleCM}        & yes                                                     & true model minus noise                                          & yes                                                                          & yes                                                            \\
\textsl{dynamic-proposed}        & yes                                                     & incremental                                          & yes                                                                          & yes                                                            \\
\textsl{dynamic-naivePrior} & yes                                                     & incremental                                          & no                                                                           & yes                                                            \\
\textsl{dynamic-fixedCM} & yes                                                     & enrollment                                           & no                                                                           & yes                                                            \\
\textsl{static}    & yes                                                     & enrollment                                           & no                                                                           & no                                                             \\
\textsl{ranked-conf}   & yes                                                     & -                                                    & -                                                                            & no                                                             \\
\textsl{linear} & -                                                       & -                                                    & -                                                                            & no                                                             \\ \bottomrule
\end{tabular}
\caption{Overview of compared settings.}
\end{table}
\label{tab:settings_overview}

Figure \ref{fig:batches} also shows several other experimental settings (see Table~\ref{tab:settings_overview} for an overview of all settings), yielding the following findings:\\
\underline{\textbf{Dynamic updates outperform static baseline:}} The setting termed \textsl{static} is a cost-sensitive baseline similar to \cite{Sperber2014}. Segmentations are optimized using \textsc{Sesla}, but not updated on-the-fly, and the cost-model is trained via enrollment. In this simulation, we assume that supervision time can be divided between enrollment and productive error correction. The longer the enrollment, the more accurate the CM, but the shorter the time for annotation. For a fair comparison, we use the same prior as in our proposed method. In the graph, we vary over how much time is reserved for enrollment, and find optimum utility for enrollments of 0 or 10 minutes, only slightly better than the confidence baseline, and 28\% worse than the best proposed setup.
Note that for large transcription projects with bigger time budgets enrollment costs are likely to amortize and this baseline can be expected to perform more strongly.\\
\underline{\textbf{Even a crude, static CM is better than no CM:}} For a comparison to cost-insensitive baselines, \textsc{Sesla} is no longer applicable, so in this experiment we consider predefined, fixed segmentations. Previous research found that transcribing (sub-) sentences is preferable over transcribing individual words \cite{Roy2009,Sperber2013}, so we divide our transcript into segments of 10 words. In \textsl{linear}, the transcriber corrects segments in linear order from the start until the time budget is exhausted, using neither CM nor UM. In contrast, \textsl{ranked-conf} uses only the UM and selects segments in order of decreasing utility until the time budget is exhausted. Deviating from the strict temporal order may be found confusing in practice, but does not affect the simulation, so results for the second baseline should be interpreted as slightly optimistic. The results show that the first baseline is much inferior, while the second baseline still slightly underperforms the proposed approach, even without updates and only the prior CM. This indicates that even using our crude prior CM is better than not using a CM at all.\\
\underline{\textbf{Updating segmentations, but not CMs, is still helpful:}} As an ablation experiment, the \textsl{dynamic-fixedCM} setting updates segmentations as proposed, but uses fixed CMs trained via enrollment as the previous baseline. This sheds light on the benefit of updating segmentations on-the-fly to reflect the actual remaining time budget at different times. Segmentation update frequency was fixed at 2.5 minutes, the fastest value employed in our simulations. Compared to \textsl{static}, this brings considerable gains, especially when setting enrollment time to 0, in which case 1540 errors are corrected (7\% worse than with the best proposed setup). This indicates that adjusting to the actual remaining time budget is a crucial factor in our framework.\\
\underline{\textbf{Naive CM prior compromises results:}} As an ablation experiment examining the importance of using the proposed prior, in \textsl{dynamic-naivePrior} we run the full proposed updating framework, but use a naive CM prior that simply assumes transcription time to take up $1s$ per word, dropping the segment overhead. This leads to considerable losses, especially when running updates infrequently, indicating that modeling segment overhead in the prior is another significant factor. \\
\underline{\textbf{Proposed updating framework almost matches oracle CM:}} As an upper bound for how much benefit can be expected from better cost modeling, in \textsl{dynamic-oracleCM} we test the full proposed updating framework, but with a CM equal to the true, oracle CM, except for the noise. That is, the CM always predicts the correct time, except for the multiplicative noise applied to the true model that captures the assumed unpredictable variations in working speed. Notice that this setup still somewhat benefits from segmentation updates because prediction errors caused by the random noise can be better recovered from. Remarkably, for high update frequencies the \textsl{dynamic-proposed} setup almost reaches the \textsl{dynamic-oracleCM} performance, with the latter gaining only 3.3\% relative. This indicates that the combination of CM- and segmentation updates is effective enough to render having a more accurate CM to start with almost unnecessary.

\subsection{Conclusions}
Based on our simulated experiments, we draw a number of first conclusions: 
\begin{itemize}
\item Our updating framework is considerably stronger than the original \textsc{Sesla} (cost-sensitive but without updates), as well as the cost-insensitive baselines.
\item The higher the updating frequency, the better the results.
\item The prior cost model can be crude, but modeling the segment overhead is important.
\item Two factors both contribute to the effectiveness of our updating framework: segmentation is updated, making sure the time budget is not over- or underspent; the cost model is updated to model the annotator more accurately.
\end{itemize}

\section{User Study}
\label{sec:user_study}
We now turn to testing our method in a realistic setting. We choose a scenario that is especially challenging in cost-sensitive annotation: we assume a new, previously unknown transcriber, who is employed to annotate only one talk. We compare the strongest dynamic (proposed) setup against the strongest static (baseline) setup, according to the simulations above. In particular, we compared the \textsl{dynamic-proposed} setting, which used the prior with overhead, updates of segmentation and CM, and a high updating frequency, against the cost-sensitive setting without dynamic updates (\textsl{static}). Enrollment was omitted, as it was deemed not worthwhile according to the simulations. Instead, we use the same prior as in the proposed system that models the overhead for each segment, but under-predicts annotation time in most cases. In practice, this leads to the annotator not being able to finish all segments chosen by the algorithm.\footnote{In contrast, consider the case of an over-predicting CM without updates: here the annotator would finish all segments before the time budget is exhausted, with the leftover time remaining unused. Our under-predicting CM is preferable as it does not result in unused annotation time.} We do not compare to cost-insensitive baselines, as earlier work indicated that cost-sensitive methods are superior \cite{Sperber2014}.

For our user experiment, we use a selection of 4 TED talks that were distinct from the ones used in the simulations (see Table~\ref{tab:data-stat}). We include 12 participants in our experiment, with English skill ranging from good to native. As user interface for transcription, we extend the \textsc{Sesla} Transcriber \cite{Sperber2014b} by implementing the necessary updating features.\footnote{Recall that updating segmentations can take up to several seconds ($\mathsection$\ref{sec:segmentation_algorithms}), but waiting time of this magnitude is undesirable. In our implementation, we start to compute an update in the background when the transcriber just started annotating a new segment. By the time the annotation of that segment is finished, the computations are usually finished, so that we can update the user interface before the annotator starts with the next segment.} Transcription is performed from scratch, but with the ASR hypothesis visible to the transcriber. This was shown to be more effective than blindly transcribing from-scratch \cite{Sperber2016}. In addition, we found it easier to automatically predict effort when transcribing from-scratch than for post-editing. We also display the ASR hypothesis of surrounding segments to the transcriber, and audio play-back includes a small portion of the preceding and succeeding segments to provide enough context for successful transcription.

\begin{table}[]
\centering
\begin{tabular}{@{}ccc@{}}
\toprule
Talk       & Reference length (\# words) & Initial ASR WER \\ \midrule
TED 1520 & 1403                   & 19.5    \\
TED 1532 & 1500                   & 11.5    \\
TED 1617 & 2175                   & 20.3    \\
TED 1685 & 1521                   & 36.7    \\ \bottomrule
\end{tabular}
\caption{Data overview.}
\label{tab:data-stat}
\end{table}

Transcribers were allowed a time budget of 20 minutes to correct as many errors as possible in a particular talk. Every transcriber worked on every talk, each time pretending it was the first piece of work for that transcriber. The 2 methods \textsl{(dynamic-proposed, static)} were alternated for each transcriber, and the assignment for which talk was transcribed with which method was shuffled between transcribers to reduce noise. Because of this randomized assignment, results are not directly comparable between different transcribers. Simply averaging over transcribers, on the other hand, would not allow to draw conclusions for individual transcribers. Instead, we use a linear mixed effects model \cite{Pinheiro2006} to explicitly account for these random factors in our experiments. We used {\it R} \cite{RCoreTeam2014} and {\it lme4} \cite{Bates2013} to perform our analysis, and tested statistical significance via the likelihood ratio test. A more detailed explanation on the employed mixed effects models is presented in \ref{sec:appendix-mem}.

\subsection{Transcriber Characteristics}
Before we turn to the evaluation of the end results, it will be helpful to analyze how transcribers differ from one another.

\begin{figure}[tb]
	\begin{center}
\includegraphics[width=0.5\linewidth]{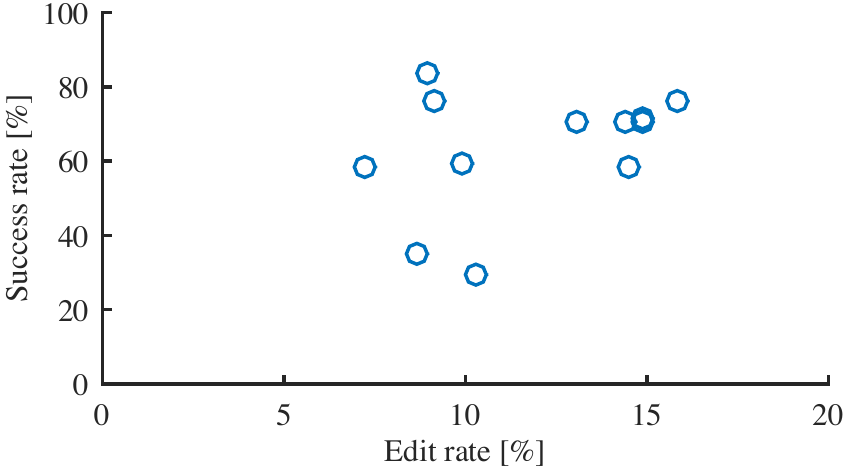}
	\end{center}
	\caption{Transcriber characteristics: {\it Edit rate} achieved within the 20-minute time budget, plotted against the {\it success rate} (proportion of edits that successfully improved the WER). According to this graph, these measures of correction speed and quality vary considerably between the 12 transcribers, and are not strongly correlated.}
	\label{fig:transcriber-stats}
\end{figure}

\underline{\textbf{Speed and accuracy varies strongly between transcribers:}} Figure~\ref{fig:transcriber-stats} shows a plot of the edit rate (edit distance between initial and corrected transcripts) achieved within the allotted 20-minute time budget as a proxy for how fast participants transcribed, and notice considerable differences between participants (edit rate between 7.2\% and 15.8\%). We also measure how many of these edits actually improved the WER as an indicator for how carefully transcribers worked, and again notice large differences (between 29\% and 85\% success rate). Quantity and quality of corrections, when measured in these terms, are only weakly correlated (Pearson correlation coefficient: 0.35).

\begin{figure}[tb]
	\begin{center}
\includegraphics[width=0.5\linewidth]{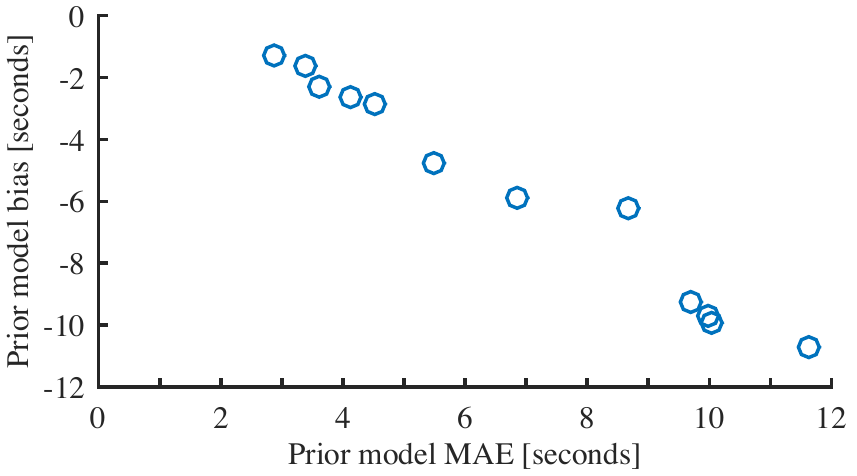}
	\end{center}
	\caption{Transcriber statistics: Mean absolute error (MAE) and bias of prior cost model on each transcriber's annotations. For all 12 transcribers, the prior model underpredicted on average (the bias is negative), although the magnitude varies strongly.}
	\label{fig:prior-accuracy}
\end{figure}

\underline{\textbf{Initial CM accuracy varies between transcribers:}} 
Figure~\ref{fig:prior-accuracy} shows the per-segment prediction accuracies of the prior CM for each participant. We computed {\it mean absolute error} (MAE) as $\frac{1}{n}\sum_{i=1\dots n}|p_i-y_i|$ and {\it bias} as $\frac{1}{n}\sum_{i=1\dots n}(p_i-y_i)$, for $n$ predicted values $p_i$ and true values $y_i$. MAE is a common regression error measure, while the bias is a measure for whether the model tends to under- or overpredict. The figure shows that the prior model underpredicted on average for every transcriber, although there were large differences in how much it underpredicted: the fastest transcriber was 1.3 seconds slower than predicted on average, the slowest transcriber 10.7 seconds slower than predicted.

The differences between transcribers we observed are a strong indicator that transcriber-specific cost modeling will be worthwhile. Note also that transcribers who differ from the prior CM more strongly have a larger potential of benefitting from the proposed dynamic CM adaptation.

\subsection{Accuracy of Cost and Utility Models}
To intrinsically assess CM performance, we compute the MAE of the predicted times given the observed times. We do this for both the initial model based on only the prior, and for trained models. For the latter, we train models for individual talks and transcribers, and perform 10-fold cross validation by training on 90\% of the observed examples and testing on the remaining 10\%, repeatedly across folds. The final number is averaged over all talks and transcribers. Results are shown in Table~\ref{tab:intrinsic_cm_um} (left). It can be seen that the trained cost models improve considerably over the initial CM, but are still not very reliable. We present more details on convergence behavior in $\mathsection$\ref{sec:dynamics}.

To evaluate the UM, we compute MAE and the Pearson linear correlation coefficient for the initially computed segments of our 4 TED talks (both selected and skipped segments). Utility predictions are normalized by segment length so that all segments have comparable influence on the intrinsic results. As shown in Table~\ref{tab:intrinsic_cm_um} (right), the UM appears reasonably strong in terms of both MAE and Pearson correlation.

\begin{table}[]
\centering
\begin{tabular}{@{}llll@{}}
\toprule
\multicolumn{2}{l}{Cost model}   & \multicolumn{2}{l}{Utility model (length-normalized)} \\ \midrule
MAE prior model              & 6.7 seconds & MAE                     & 0.19    \\
MAE 10-fold cross validation & 4.9 seconds & Pearson correlation     & 0.57    \\ \bottomrule
\end{tabular}
\caption{Intrinsic CM and UM performance.}
\label{tab:intrinsic_cm_um}
\end{table}

\subsection{Main Results: Transcription Efficiency}

\begin{figure}[tb]
	\begin{center}
\includegraphics[width=0.5\linewidth]{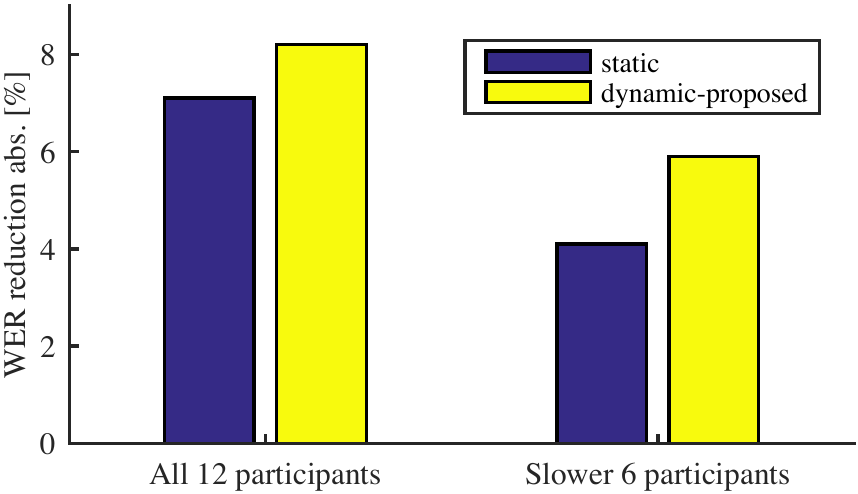}
	\end{center}
	\caption{WER reduction for the two evaluated methods and different groups of participants.}
	\label{fig:gains-rel}
\end{figure}

\underline{\textbf{Proposed method improves overall efficiency:}} For an empirical evaluation of the end results, we fit a simple mixed effects model. The model predicts the WER improvement, based only on a binary input that indicates whether \textsl{dynamic-proposed} or \textsl{static} was used. The particular transcriber and talk are incorporated in the model as random effects. Figure~\ref{fig:gains-rel}~(left) shows the result: the proposed method reduced the WER by 8.2\% absolute, the baseline only by 7.1\% absolute, making the baseline 15\% more efficient in terms of corrected errors per time. However, statistical significance of the used method being responsible for the difference is quite weak (p=0.18).

\underline{\textbf{Transcribers deviating from prior CM benefit most:}} To analyze further in what situations we can expect reliable gains from using the proposed method, we estimate the impact of the used method on a per-transcriber basis. This is possible by fitting a refined model with an additional random effect that models how strongly each transcriber is affected by what method was used. The results reveal that the gains were especially strong for the slower transcribers (who deviated from the prior model the most), and much smaller for the faster transcribers. In fact, the Spearman rank correlation between prior CM bias (Figure~\ref{fig:prior-accuracy}) and observed gain is a highly positive value at 0.78. This can be explained by observing that faster transcribers, who were more accurately modeled by the initial CM already, have little room for improvement from adapting the CMs. The slower transcribers, on the other hand, deviated strongly from the initial CM, and benefitted much more from the dynamic updating framework. Figure~\ref{fig:gains-rel}~(right) illustrates the fit of the original model when using only the slower 50\% of participants (which coincided with the 50\% of participants with largest negative prior model bias). Differences are much more pronounced: 4.1\% WER reduction for the baseline, 5.9\% for the proposed method, a 42\% relative productivity increase. The difference is statistically significant (p=0.027) for this subgroup of users.

\begin{figure}
\begin{subfigure}{.5\textwidth}
  \centering
  \includegraphics[width=.8\linewidth]{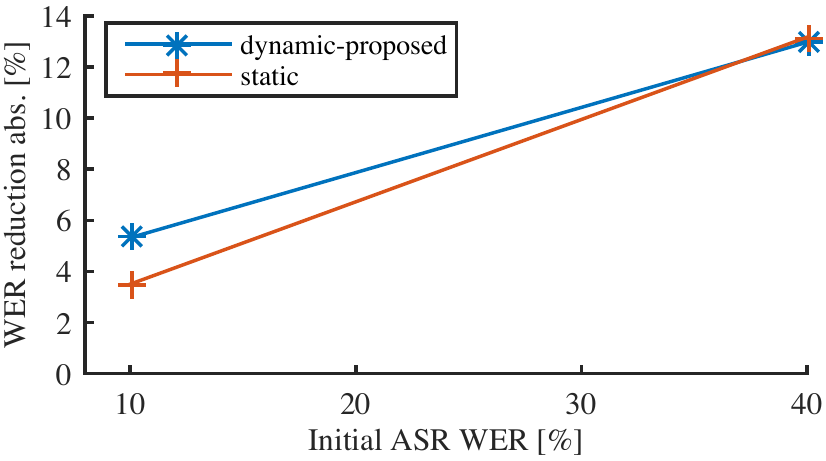}
  \caption{All 12 participants}
  \label{fig:gains_vs_asr_all}
\end{subfigure}%
\begin{subfigure}{.5\textwidth}
  \centering
  \includegraphics[width=.8\linewidth]{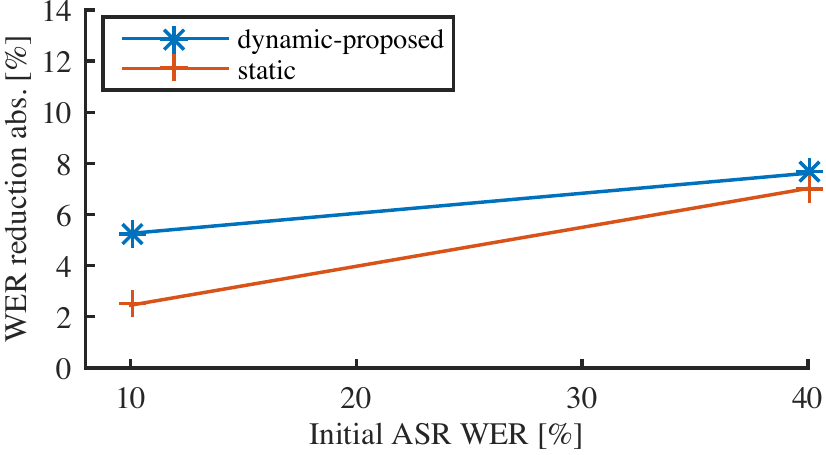}
  \caption{Slower 6 participants}
  \label{fig:gains_vs_asr_slower}
\end{subfigure}
\caption{WER reduction plotted against the initial ASR WER, for the two evaluated methods and different groups of participants.}
\label{fig:gains_vs_asr}
\end{figure}

\underline{\textbf{Gains are strongest when ASR WER is low:}} We fit another mixed effects model that includes the initial ASR WER as a fixed effect. The higher the initial ASR WER, the more errors can potentially be removed. We model the influence of the initial ASR WER separately for baseline and proposed method. The results are visualized in Figure~\ref{fig:gains_vs_asr_all}. First, note that as expected, the initial ASR WER has a strong influence, with more errors being corrected for initially bad transcripts. Moreover, we confirm that in general our proposed method reduces the WER more strongly than the baseline method with this more flexible model, as well. A perhaps surprising observation is that transcripts with low initial WER seem to benefit most from the proposed method. We hypothesize that this can be explained as follows: When the initial WER is high, even randomly selected segments have a relatively high chance of containing a high proportion of errors that can be corrected. For low initial WERs, on the other hand, it is more important to pick carefully what segments to spend time on, because many candidate segments contain only few errors. In the latter case, the more accurate CMs evidently help making an appropriate selection of segments. Figure~\ref{fig:gains_vs_asr_slower} shows the same model again for only the slower half of participants, again revealing a much larger gain from the proposed method. Statistical significance is again stronger: p=0.084 for the subgroup as opposed to p=0.33 for all participants.

\subsection{Analysis of Dynamics}
\label{sec:dynamics}

In the previous section, we confirmed that our method improves overall correction efficiency. Now, we take a closer look at the dynamic behavior of our updating framework.

\begin{figure}[tb]
	\begin{center}
\includegraphics[width=0.5\linewidth]{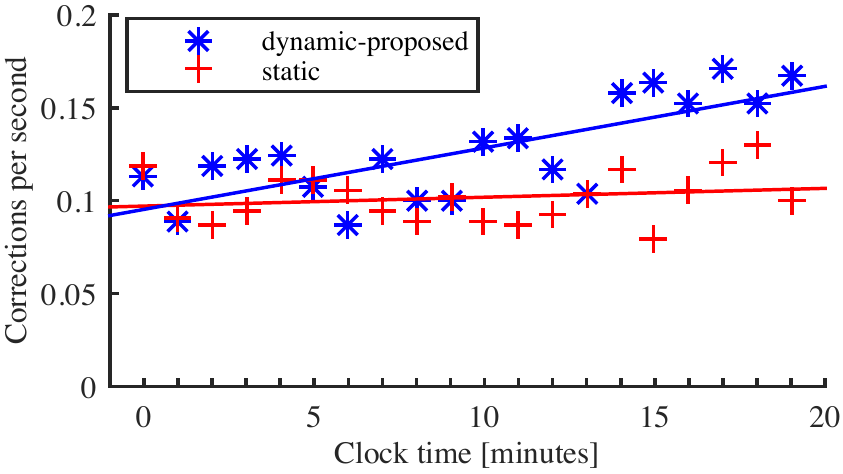}
	\end{center}
	\caption{Efficiency over time, averaged over all transcribers and talks, and grouped into bins for each minute on the clock. The efficiency of \textsl{dynamic-proposed} visibly improves over time, as cost models are updated, while \textsl{static} efficiency remains constant.}
	\label{fig:efficiency-over-time}
\end{figure}

\underline{\textbf{Transcription efficiency improves over time:}} We first analyze how efficiency changes as CMs are improved. We compute efficiency for each corrected segment as number of corrected errors in the segment divided by editing time. We expect that as the CMs improve during transcription of a talk, so will the efficiency. Figure~\ref{fig:efficiency-over-time} demonstrates that this is the case: the further the clock time progresses within the given 20 minute time budget, the higher the per-segment efficiency for the proposed method. In this graph, results for all transcribers are averaged. The static baseline does not change over time, as expected. Note also for the proposed method that efficiency does not appear to have converged after the 20 minutes worth of transcription time.

\begin{figure}
\begin{subfigure}{.33\textwidth}
  \centering
  \includegraphics[width=.9\linewidth]{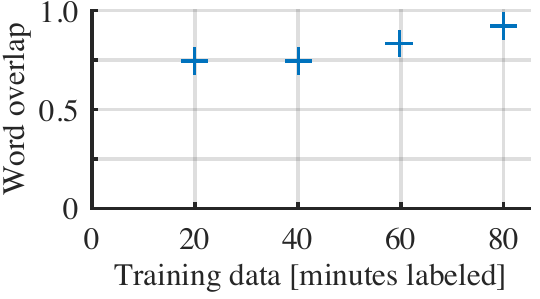}
  \caption{All segments shuffled.}
  \label{fig:cm_convergence_shuff}
\end{subfigure}%
\begin{subfigure}{.33\textwidth}
  \centering
  \includegraphics[width=.9\linewidth]{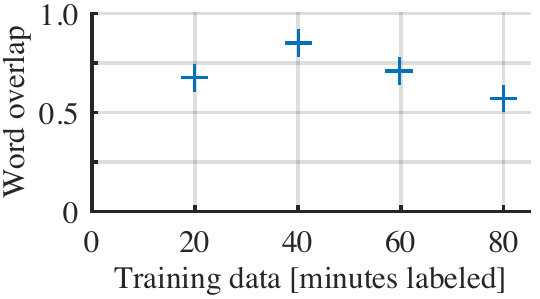}
  \caption{Adding complete talks.}
  \label{fig:cm_convergence_full}
\end{subfigure}
\begin{subfigure}{.33\textwidth}
  \centering
  \includegraphics[width=.9\linewidth]{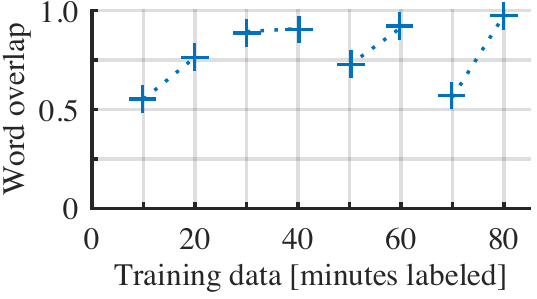}
  \caption{Adding half talks.}
  \label{fig:cm_convergence_half}
\end{subfigure}
\caption{Cost model convergence, indicated by proportion of overlapping words (y-axis) between consecutive re-segmentation of a talk by using more and more CM training examples (x-axis). When shuffling segments from all talks, cost models seem to converge (\ref{fig:cm_convergence_shuff}). When instead adding all annotations from a particular talk at once, cost models seem to strongly diverge from the previous model (\ref{fig:cm_convergence_full}). When adding annotations from half talks, adding the first half of a talk causes the model to diverge from the previous model, while after adding the second half the model remains more similar (\ref{fig:cm_convergence_half}).}
\label{fig:cm_convergence}
\end{figure}

\underline{\textbf{CMs converge within but not across talks:}} To analyze long-term CM convergence, we use the logged annotation times for all 4 talks transcribed per transcriber. We shuffle the data points, and train CMs for each transcriber by progressively adding more data to the training. We measure convergence by feeding the trained CMs into \textsc{Sesla} to select segments from a fixed, held-out TED talk. We measure how many words overlap between segmentations created using different CMs: we expect that when the CM has converged, adding more training data to it will not change the segmentation significantly, leading to a word overlap close to 1. Figure~\ref{fig:cm_convergence_shuff} shows the averaged convergence over all participants. The figure suggests that CMs are almost converged after the 80 minutes of annotation. However, this analysis does not show the full picture, as becomes evident when repeating the same experiment but without shuffling the data. That is, we first add all the annotation data from the first talk to the training, then the second talk, and so on. Figure~\ref{fig:cm_convergence_full} shows the result, and reveals that in fact CMs are diverging across talks. Apparently the transcribers' working speeds are relatively consistent within a transcribed talk, but vary strongly between different talks. Potential causes could be factors such as familiarity with topic and speaking style, fatigue, etc. Figure~\ref{fig:cm_convergence_half} provides some more evidence to support this explanation. Here, we add training data not for the whole talk, but progressively for half talks, keeping the same order. We can see that in general, when adding the first half of annotated data of a talk to the CM training, the segmentation overlap tends to be quite low, similar to the previous experiment. However, when adding the second half, the overlap is much bigger. This again indicates that CM training data is relatively consistent when the annotations stem from the same talk, but much less consistent when moving to a new annotated talk. We conclude from these observations that task-dependent differences may dominate convergence, which would be another strong reason to prefer our proposed updating approach over a static, enrollment-based approach that cannot adapt to these observed dynamics.

\subsection{Discussion}
We conclude from our experiment that our updating framework is effective in the challenging scenario investigated. Gains over the static baseline were particularly strong when a transcriber deviated strongly from the prior CM, and when initial ASR transcript was already of relatively high quality. Importantly, we saw that transcribers have very different working speeds, making a strong case for transcriber-dependent modeling. Moreover, the observation that transcription effort varies depending on the particular data indicates that CMs learned on-the-fly are a better choice than static CMS (transcriber-dependent or not), even if enrollment effort is amortized over time.

\section{Conclusion}
We proposed a new dynamic updating framework for cost-sensitive correction of speech transcripts, along with a new, faster algorithm that determines which segments to correct. Besides being more convenient to use than previous cost-sensitive methods that required transcriber enrollment, we show in a simulated study that our approach yields higher supervision efficiency. Efficiency gains are attributed to updating cost models, updating segmentations, and a sensible initial cost model.

We ran a user study for the scenario where every transcriber conducts only a small amount of work, which is especially challenging in the context of cost-sensitive annotation. We observed productivity gains of 15\% on average across all 12 participants, and 42\% for the 6 participants whose working style deviated the most from the initial cost model. Besides confirming the effectiveness of our proposed updating framework, the large gap between different transcribers indicates that transcriber-specific cost modeling is crucial.

For future research, we suggest investigating how dynamically updating cost models can optimally capture the per-transcriber variance in productivity caused by influences such as fatigue and topic familiarity. Extension to other tasks, especially post-editing for machine translation, seems promising. A further scenario worth investigating is multi-pass transcription \cite{Stuker2012} with high quality requirements, where the first pass could be made cheaper using our proposed method. Moreover, additional passes using our method could be allowed if more time budget becomes available at a later point. Finally, we suggest investigating dynamic updates of a transcriber dependent utility model.

\section*{Acknowledgments}
We thank the anonymous reviewers for their valuable comments. We also thank Oliver Adams, Philip Arthur, Tuna Murat Cicek, Angela Grimminger, Michael Heck, Matthew Holland, Joseph Irwin, Nurul Lubis, Santiago Escobar Martinez, Patrick Lumban Tobing, and Micha Wetzel for their help with the user experiments.

\appendix

\section{Mixed Effects Models}
\label{sec:appendix-mem}

We give a brief introduction to mixed effects models, and refer to the relevant literature for details \cite{Pinheiro2006}. Mixed effects models are a tool to analyze data that is influenced by ``random" factors that are difficult to control for, and that potentially have a significant influence on our observations. In our setting in particular, we are concerned about the influence of particular transcribers and particular talks. Many characteristics of the particular transcriber and talk can influence transcription speed and effectiveness. Because it is virtually impossible to account for these influences, we consider them as noise that we wish to get rid of. In fact, this is a common problem in user studies. Recently, mixed effects models have become a popular a way of dealing with such situations \cite{Goldwater2010,Federico2014,Green2013}.

Mixed effects models are specified by the following components:
\begin{itemize}[leftmargin=*]
\item {\it Response variable}: The central quantity for which we wish to determine how it is influenced by other measured covariates. In our experiments, this is the reduction in word error rate achieved in a particular transcription run.
\item {\it Fixed effects}: Numerical or categorical attributes that influence the response variable in a meaningful way. In this paper, we assume a linear relationship. Our fixed effects are (1) the user interface, and (2) the initial ASR WER which has a strong influence on how many errors are corrected and should potentially be explicitly accounted for.
\item {\it Random effects}: Categorical factors that are hard to control for or hard to understand. Generally, the observations include only a limited sample of values out of a large set of possible values (as is the case with the particular transcribers and talks that are part of our experiment). For each random effect an intercept is estimated by the model. Thus, e.g.\ for each transcriber a mean WER reduction is estimated to explain some of the observed variance. Random effects are modeled to follow a Gaussian distribution with the observed sample mean and variance to be estimated.
\item {\it Error term}: The variance in observations that is not explained by the random effects is finally modeled by the general error term.
\end{itemize}

For the simple models in our experiments, we restrict ourselves to {\it linear} mixed effects models. These can be seen as an extension of linear regression models, extended by the random effects component. We tested the significance of the fixed effects in our models via likelihood ratio test. That is, we built a null-model with the effect in question removed, and examined whether this significantly reduced the model likelihood.



\bibliographystyle{elsarticle-num}

\bibliography{library}

\end{document}